\documentclass{article}
\usepackage{ijcai19}

\pdfpageattr{/Group <</S /Transparency /I true /CS /DeviceRGB>>}
\usepackage{times}
\usepackage{url}
\usepackage[hidelinks]{hyperref}
\usepackage[utf8]{inputenc}
\usepackage[small]{caption}
\usepackage{graphicx}
\usepackage{amsmath}

\usepackage{booktabs}
\usepackage{algorithm}
\usepackage{algorithmic}

\usepackage{esvect}
\usepackage{mathtools}
\usepackage{amssymb}
\usepackage{stmaryrd}

\DeclareMathOperator*{\argmax}{arg\,max}

\makeatletter
\newcommand{\xmapsfrom}[2][]{%
\ext@arrow3095\leftarrowfill@{#1}{#2}\mapsfromchar
}
\makeatother

\newtheorem{algo}{Algorithm}{\bf}{\it}

\title{Approximate Dynamic Programming with Neural\\ Networks in Linear Discrete Action Spaces}

\author{
Wouter van Heeswijk$^1$\And%\and
Han La Poutr{\'e}$^{1,2}$%\And
\affiliations
$^1$CWI, Amsterdam, The Netherlands
$^2$TU Delft, Delft, The Netherlands\\
\emails
\{Wouter.Van.Heeswijk, Han.La.Poutre\}@cwi.nl
}

\begin{document}

\maketitle

\begin{abstract}
 Real-world problems of operations research are typically high-dimensional and combinatorial. Linear programs are generally used to formulate and efficiently solve these large decision problems. However, in multi-period decision problems, we must often compute expected downstream values corresponding to current decisions. When applying stochastic methods to approximate these values, linear programs become restrictive for designing value function approximations (VFAs). In particular, the manual design of a polynomial VFA is challenging. 
This paper presents an integrated approach for complex optimization problems, focusing on applications in the domain of operations research. It develops a hybrid solution method that combines linear programming and neural networks as part of approximate dynamic programming.
Our proposed solution method embeds neural network VFAs into linear decision problems, combining the nonlinear expressive power of neural networks with the efficiency of solving linear programs. As a proof of concept, we perform numerical experiments on a transportation problem. The neural network VFAs consistently outperform polynomial VFAs, with limited design and tuning effort.
\end{abstract}

\section{Introduction}
Problems in operations research (OR) are generally concerned with allocating resources, aiming to maximize some reward function. Applications of OR are found in domains such as transportation, energy, and manufacturing. Although many effective solutions -- particularly linear programs (LPs) -- exist for static problems, solving dynamic problems over a time horizon remains challenging, as we need downstream values corresponding to current decisions. Estimating these values is often difficult within linear programming settings.

We address the integration of neural networks and linear programs in the context of Approximate Dynamic Programming (ADP). Spurred by increasing availability of both data and computing power, neural networks are successfully applied in many fields. Their potential applications have also been identified for ADP, yet their use is not widespread. This paper extends an effort in this direction, explicitly considering implementation for problems with large action spaces.

A key challenge of ADP is to reliably estimate the downstream value corresponding to actions, enabling to learn a policy that maximizes value over the full planning horizon. One research stream within ADP focuses on value function approximations (VFAs) to estimate downstream values. In this approach, we design a set of features (explanatory variables) and organize them as a polynomial function that represents the value of being in a given state and perform linear regression to learn weights associated to the features. Although polynomial VFAs often yield satisfactory results, designing the features is challenging. Particularly higher-level interactions are difficult to grasp for human designers.

The problem of features design is amplified by the large action spaces that are encountered in typical ADP settings. It is common for the action space growing too large to enumerate within reasonable time. Formulating the decision problem as a mathematical program -- preferably a linear program for efficient solving -- preserves optimality and often vastly enhances the magnitude of problems that can be handled. Even if the action space is not overly large, there may be reasons to use mathematical programming, such as higher speed or the availability of existing model formulations. However, mathematical programming poses additional challenges for VFAs. Assuming a linear program, the variables representing the features in the objective function must be linear as well. Although nonlinear features might be designed, they must be expressed as linear systems, often requiring complicated constructions of artificial variables. Complex polynomial VFAs are therefore difficult to embed in LPs.

The integration of neural networks within LPs for decision making addresses this challenge. Neural networks are able to learn complex nonlinear functions; theoretically, a single-layer network may learn any continuous function \cite{cybenko1989}. Because neural networks are not restricted by linearity, they may identify nonlinear structures between lower-level features, without explicitly defining these as features within the LP. The activation functions should be transformable into piecewise linear functions. Fortunately, most  modern neural networks satisfy this condition. 

In this paper, we design a hybrid approach to address ADP problems based on neural networks and LPs. The planning problem tested is a dynamic transport problem inspired by practice. To preserve focus on the methodological aspect of the paper, we will not discuss design choices in detail. The problem serves as a test case that is sufficiently rich and challenging to adequately test the solution method.

This paper contributes to the state of the art in the following ways. First, we design a hybrid approach for integrating neural networks and LPs to tackle ADP problems. Second, we  provide insights into the performance of various neural network structures based on numerical experiments, specifically the quality and computational effort. Third, we show that neural network VFAs significantly improve upon current practice, which is based on polynomial VFAs.

\section{Related work}
%Large action spaces
%Compared to the research efforts made in approximating states and value functions, studies that focus on controlling the action space has been relatively limited \citet{dulac2015}.\citet{pazis2011} propose to partition the action space into subsets, learning a subpolicy for each subset. The amount of information that needs to be stored in their approach grows linearly with the dimension of the action, whereas the true action space itself typically grows exponentially. Both of them integrate their approaches in linear programs. \citet{dulac2012} have a comparable partitioning approach.\citet{dulac2015} propose a neighborhood search heuristic to deal with large action spaces. 
%Estimating the weights corresponding to features in ADP methods typically takes place using some form of least-squares regression; \citet{powell2011} presents a number of these approaches. 

Given the successful applications of neural networks in regression, their application on ADP problems seems natural. The idea is not novel; the seminal work of \cite{bertsekas1995} already presents the use of feature vectors as input to neural networks as an established concept. Also \cite{powell2011} describe neural networks as a powerful tool for ADP algorithms. However, neural network VFAs have not yet been well-tested for large action spaces -- where the neural network cannot be used to enumerate the downstream value for every action -- we are not aware of previous studies addressing the integration of neural networks into decision-making LPs for this type of problems. We highlight some relevant works in ADP and reinforcement learning, discussing the most closely related applications of neural networks and linear programming.

We start with neural networks in ADP. \cite{bertsekas2008} discusses applying neural networks in ADP, coining the term neuro-dynamic programming. He broadly defines neural networks as essentially nonlinear VFAs, using either the full state or a smaller feature vector as input. Alternatively, neural networks may also be used as a pre-processing step to extract feature vectors from the state. According to \cite{powell2011}, neural network VFAs have mainly been applied on classical engineering problems that typically have low-dimensional action spaces. \cite{schmidhuber2015} provides an survey of deep learning studies, including the use of neural networks in reinforcement learning. The neural networks are generally used to learn values associated to state-action pairs, i.e., as VFAs. No mention is made of embedding such VFAs in linear programs. \cite{vanheeswijk2018} study shallow neural network VFAs in a transportation context, but require full enumeration of the action space.

We proceed to discuss linear programming in ADP. \cite{defarias2003} study the linear programming approach for ADP, assuming linearly defined VFAs. \cite{powell2016} states that decision problems with tens of thousands dimensions can generally be solved with modern commercial solvers. However, when instances become too vast, also linear programming may require unsatisfactory computational times. \cite{dulac2012} and \cite{pazis2011} propose factorization methods to divide the action space into linear subproblems, exponentially reducing the computational effort. The size of the state space is a limiting influence in their solution. In the transportation domain, \cite{perez2017} and \cite{vanheeswijk2019} provide recent examples of polynomial VFAs integrated in linear programs.

\section{Solution method}
We briefly introduce the notation for Markov decision problems (MDPs) as used in this paper. MDP models are useful to mathematically model decision problems with stochastic and dynamic properties. In OR, many are combinatorial optimization problems. An MDP is a stochastic control process for which the objective is to maximize rewards (or minimize costs) over a discrete time horizon $\mathcal{T}$, with decision epochs $t\in \mathcal{T}$ separated by equidistant time intervals. A discounted MDP can be described by $( \mathcal{S} , \mathcal{X}(S) , \mathbb{P}(S^\prime|S,x) , R(S,x) ,\rho)$, with $\mathcal{S}$ being the set of problem states, $\mathcal{X}(S)$ being the set of feasible actions when in state $S\in \mathcal{S}$, $\mathbb{P}(S^\prime|S,x)$ being the transition probability of transitioning from state $S$ to $S^\prime \in \mathcal{S}^\prime \subseteq\mathcal{S}$ after taking action $x \in \mathcal{X}(S)$, $R(S,x)$ being the direct reward when taking action $x$ in state $S$, and  $\rho \in [0,1)$ is the discount rate applied to future rewards. The Bellman equation yields the maximum value corresponding to each state:
\begin{align} 
V(S)=\max_{x\in\mathcal{X}(S)} \bigg(R(S,x)+ \rho \smashoperator{\sum_{S^\prime\in \mathcal{S}^\prime}} \mathbb{P}\left(S^\prime|S,x\right) V(S^\prime)\bigg)\enspace.\notag
\end{align}

%\begin{align} 
%V(S)=\max_{x\in\mathcal{X}(S)} \bigg(R(x)+ \gamma \smashoperator{\sum_{\omega\in\Omega}} \mathbb{P}\left(W(\omega)\right) V(S^\prime|S,x,\omega)\bigg)\enspace,\forall S \in \mathcal{S} \enspace.
%\end{align}

Solving the Bellman equation for all states yields the optimal policy. Several techniques exist to accomplish this, yet for many realistic problems these are computationally intractable. The next section addresses this issue.

\subsection{Approximate Dynamic Programming}
Approximate Dynamic Programming (ADP) is a framework to learn policies for MDPs that are too large to solve exactly within reasonable time. This section provides a short and high-level overview. We refer to \cite{powell2011} for an extensive discussion on the topic. At its core, ADP uses Monte Carlo simulation to sample rewards and estimate the downstream values of state-action pairs, enabling to learn good policies without exhaustively exploring the MDP.

From a computational perspective, problems may arise in three areas of MDPs, namely the sizes of the state space (number of states), action space (number of actions per state) and outcome space (number of possible outcomes per action). Multiple solution approaches exist for each of these areas; we restrict ourselves to the ones used in this paper.

We start with the outcome space $\mathcal{S}^\prime \subseteq \mathcal{S}$. To identify the best action in any state, the Bellman equation requires computing $V(S^\prime)$ for each $S^\prime \in \mathcal{S}^\prime$, where $\mathcal{S}^\prime$ might be unique for each state-action pair. ADP circumvents this procedure by instead attaching a single value to a state-action pair. Thus, we replace the stochastic expression
$\smashoperator{\sum_{S^\prime\in \mathcal{S}^\prime}} \mathbb{P}\left(S^\prime|S,x\right) V(S^\prime)$ with a deterministic value function $V(S,x)$. For each state-action pair, we only need to evaluate one downstream value rather than $|\mathcal{S}^\prime|$ outcomes. This downstream value is estimated by repeated Monte Carlo sampling, i.e., we randomly draw outcome states $S^\prime$ and observe their values.

Next, we discuss the state space $\mathcal{S}$. In many optimization problems the state is a high-dimensional vector with numerous possible realizations. Computing the value for each individual state may therefore be intractable. Therefore, we replace the true value function with a value function approximation (VFA) $\bar{V}(S,x)$. The VFA is a function that returns an expected value given a set of features (explanatory variables) that capture the essential information in state-action pairs needed to estimate their value. The VFA design is further discussed in Section~\ref{ssec:vfa}.

Finally, we address the action space $\mathcal{X}$. In combinatorial problems, this space quickly grows beyond the limits of enumeration. As we need thousands of observations to learn a good policy, each decision problem should typically be solvable within a few seconds. To avoid enumerating the full action space, the decision problem may be expressed as a mathematical program. In particular LPs are well-studied; modern solvers often solve such problems highly efficiently. Mathematical programs can be solved to optimality, while significantly upscaling the action space sizes that can be handled. 
%Second, heuristics may be designed to identify high-quality candidate solutions. Although they do not guarantee preserving the 'optimal' action -- given the value approximations that are already made -- a well-designed heuristic is able to quickly generate good solutions. 

The outline of the ADP algorithm is now presented. We use $N$ iterations to learn the VFA; each iteration represents a discrete time step. At every iteration $n$, the action maximizes expected value given the prevailing VFA $\bar{V}_{n-1}(\cdot)$, resulting in the following observed value:
\begin{align}
\hat{v}_n=\max_{x_n\in\mathcal{X}(S_n)} \bigg(R(S_n,x_n)+ \rho \bar{V}_{n-1}(S_n,x_n)\bigg) \enspace. \notag
\end{align}

The difference between expected value for the preceding state-action pair at $n-1$ (i.e., $\bar{V}_{n-1}(S_{n-1},x_{n-1})$) and the observation at $n$ (i.e., $\hat{v}^n$) updates the VFA, using an updating function $\bar{V}_{n}(\cdot)\mapsfrom U\big(\bar{V}_{n-1}(\cdot),S_{n-1},x_{n-1},\hat{v}_{n}\big)$. Algorithm 1 shows the outline of the ADP algorithm to learn the VFA.

\setcounter{algo}{0}
\begin{algo}\label{ADPalgorithm}Basic ADP algorithm to learn the VFA. 
\end{algo}
\footnotesize
%\vspace{2mm}
\begin{tabular}{ l  l }
\toprule			
1: &  \textbf{initialize} $\bar{V}_{0}(\cdot)$\\
2: & $n \mapsfrom 1$ \\
3: & $S_{1} \xmapsfrom{\mathbb{P}}\mathcal{S}$ \\
4: & \textbf{while} $n \leq N$ \textbf{do} \\					
5: & \qquad $x_{n} \mapsfrom  \argmax\limits_{x_{n}\in\mathcal{X}(S_{n})} \bigg(R(S_n,x_n)+\rho \bar{V}_{n-1}(S_{n},x_{n})\bigg)$   \\
6: & \qquad$\hat{v}_{n} \mapsfrom \bigg(R(S_n,x_n)+\rho \bar{V}_{n-1}(S_{n},x_{n})\bigg)$\\
7: & \qquad $\bar{V}_{n}(\cdot)\mapsfrom U\big(\bar{V}_{n-1}(\cdot),S_{n-1},x_{n-1},\hat{v}_{n}\big)$ \\
8: & \qquad $\mathcal{S}^\prime\mapsfrom (S_n,x_n)$ \\
9: & \qquad $S_{n+1}\xmapsfrom{\mathbb{P}} \mathcal{S}^\prime$ \\		
10: & \qquad $n\mapsfrom n+1$ \\
11: & \textbf{end while} \\
12: & \textbf{return} $\bar{V}_{N}(\cdot)$  \\
\bottomrule  
\end{tabular}
\normalsize
\vspace{4mm}

\subsection{Polynomial VFA (PL-VFA)}\label{ssec:vfa}
This section addresses the VFA in more detail. As mentioned earlier, we operate on features that are extracted from state-action pairs. Let $\mathcal{F}$ be the set of indicators describing the features, with each indicator $f \in \mathcal{F}$ referring to some representative feature of a state-action pair. We define a contractive mapping $\phi$ that extracts features for any given state-action pair, i.e.,  $\phi:(S,x)\mapsto \mathbb{R}^{|\mathcal{F}|}$, the corresponding vector of features is $[\phi_f]_{\forall f \in \mathcal{F}}$. Formally, the VFA is described by $(\bar{V} \circ \phi):\mathcal{S}\times\mathcal{X}\mapsto \mathbb{R}$.

VFAs are commonly designed in polynomial form (PL-VFA). Let $w_f \in \mathbb{R}$ be a weight associated to feature $\phi_f \in \mathbb{R}$. Then, the polynomial VFA may be described by $\bar{V}(S,x)=\sum_{f\in\mathcal{F}} w_f \phi_f(S,x)$. PL-VFAs are popular for several reasons. Polynomials are able to approximate most functions, an appropriate polynomial  in theory approaches the true value function arbitrarily close. Furthermore, although the features may be nonlinear, the expression itself is linear. It can therefore be incorporated into linear programming formulations. Techniques such as temporal-difference learning may be used to update the weights \cite{sutton2018}.

Although polynomials might theoretically approximate the true value function, randomly defining a polynomial will likely not perform well \cite{powell2016}. A properly designed PL-VFA is aligned with the structure of the value function. This manual design of VFAs is a key challenge for successful implementations, requiring careful modeling and testing of individual value functions. This is where the linear formulation becomes restrictive, as features representing higher-order effects must be explicitly modeled. Additional problems arise when we resort to linear programming to handle large action spaces. It then becomes challenging to express non-linear features in linear form. Such conversions often require complicated structures involving many artificial variables. 

To overcome the limitations of polynomial VFAs, the VFA may be expressed by neural networks. The nonlinear architecture of such networks allows to unravel complex structures, even when inputs are linear operands of state-action pairs. We further discuss neural network VFAs in the next section.

\subsection{Neural network VFA (NN-VFA)}
A general introduction to neural networks is provided by \cite{gurney2014}, we only address the VFA design. In neural network VFAs (NN-VFAs), the feature vector $[\phi_f]_{\forall f \in \mathcal{F}}$ is transformed by a weighted set of nonlinear activation functions (neurons), resulting into a single output value $\bar{V}(S,x)$. Compared to the PL-VFA, the main advantage is that the NN-VFA may learn higher-order effects that are not explicitly defined in the feature vector. We emphasize that the input quality remains crucial for the NN-VFA performance, but feature design is comparatively easier than for PL-VFAs.

The NN-VFA is composed of an input layer (the feature vector), a least one hidden layer containing neurons, and an output layer with a single node that returns the expected value for the given state-action pair \cite{vanheeswijk2018}. In a fully connected network, every neuron in the network connects to all neurons in the preceding layer. Each neuron receives the inner product of all neurons in the preceding layer and their corresponding output weights as input and transforms it into a single neuron value. 

The NN-VFA contains $K\geq 1$ hidden layers; we use $\mathcal{K}\triangleq\{1,\ldots,K\}$ to denote the set of hidden layers. The indicator $k=0$ refers to the input layer that contains the features; layer $K+1$ is the output layer. Furthermore, the index $d_k\in\mathbb{N}$ refers to a specific neuron in layer $k\in \mathcal{K}$, with $\mathcal{D}_k$ denoting the set of neurons in layer $k$. Each neuron represents a nonlinear activation function $\sigma_{d_k}$, corresponding to neuron $d$ in layer $k$. For layers $k>0$, an input weight $w_{d_k,d_{k-1}}\in \mathbb{R}$ describes the weight of a neuron as input for $d_k$; the vector $\vv{w}_{d_k}=[w_{d_k,d_{k-1}}]_{\forall d_{k-1}\in \mathcal{D}_{k-1}}$ denotes all inbound weights for neuron $d_k$.

We introduce some additional notation to describe the neuron values. The value of neuron $d$ in layer $k$ is described by $y_{d_k}$; the value vector for layer $k$ is given by $\vv{y}_{k}=[y_{d_k}]_{\forall d_k \in \mathcal{D}_k}$. The input layer equals the features, i.e.,  $\vv{y}_0=[y_{d_0}]_{\forall d_{0}\in \mathcal{D}_0}\triangleq[\phi_f]_{\forall f \in \mathcal{F}}$, with $|\mathcal{D}_0|=|\mathcal{F}|$. The values of the neurons are expressed by $y_{d_k}\triangleq\sigma_{d_k}(\langle \vv{y}_{k-1},\vv{w}_{d_k}\rangle)$. Finally, the output value of the network is given by $\bar{V}(S_{n-1},x_{n-1})\triangleq y_{d_{K+1}}=\langle \vv{y}_{K},\vv{w}_{d_{K+1}}\rangle$.

Activation functions in neural networks are nonlinear. Therefore, they cannot be directly computed within linear programs. However, most common activation functions in contemporary neural networks can be modeled by simple piecewise linear functions. Integration of the NN-VFA in linear programs is discussed in the next section.

\subsection{Integrating the NN-VFA in LPs}
Nowadays, many neural networks use (variants of) rectified linear units (ReLUs) as activation functions \cite{wilmanski2016}. A ReLU returns either 0 or its input value, whichever is larger. They can be represented by a piecewise linear function with two components, allowing to incorporate them in the LP designed to solve the decision problem. Each state-action pair has a unique expected downstream value. To evaluate actions, the neural network must therefore be expressed as a set of linear equations. We follow an implementation comparable to that of \cite{bunel2018}, using binary variables and big M constraints to correctly compute the ReLU values. Additional artificial variables are required to compute the basis functions corresponding to actions. To preserve linearity of the action problem, the features should be linear expressions that can be derived from $[S,x]$.

We use a stochastic gradient descent (SGD) algorithm to update the weights, meaning that the network weights are iteratively adjusted after each iteration and corresponding observation $\hat{v}_{n}$ \cite{haykin2009}. At $n=0$, we use He initialization to generate starting values for the weights \cite{he2015}.  The learning rate $\eta \in (0,1]$ determines how responsive the weights are to observations deviating from the estimate.

\section{Experimental design}
To validate the solution method as well as the performance of the NN-VFA, we run a number of numerical experiments that compare it to the PL-VFA. We evaluate both the behavior of the VFAs $\bar{V}_N(\cdot)$  under varying circumstances and the performance $R(\cdot)$ of the resulting policies.

For a clear comparison that distills the essential insights, we keep the applications basic. To update the weights, we use TD(0) for the PL-VFA  and SGD  for the NN-VFAs, always using the same learning rate $\eta$.  In all cases, we use He initialization to set the weights at $n=0$. We use pure exploration to acquire value observations, i.e., each decision maximizes the expected value given the prevailing policy. Furthermore, we deliberately do not put excessive effort into design and fine-tuning; the main goal of the NN-VFA is to reduce the manual design effort compared to the PL-VFA.
%We note that TD(0) and SGD are equivalent for networks with 0 hidden layers.

The experiments compare a PL-VFA to two neural network VFAs: the NN(1,20)-VFA (1 layer, 20 neurons) and the NN(3,20)-VFA (3 layers, 20 neurons per layer). Although a single-layer network theoretically suffices to learn a function, deep neural networks may model the same function with significantly fewer neurons \cite{delalleau2011}. In fact, for many common functions, the required number of neurons decreases exponentially with the number of layers \cite{lin2017}. \cite{rolnick2018} suggest that, for many functions encountered in practical settings, relatively small networks suffice to accurately describe functions. Downsides of deeper networks are the longer training time and potential loss of information \cite{huang2016}.

The experimental design is as follows. First, we compare convergence properties of VFAs. Second, we perform experiments on various neural network configurations and learning rates,  giving insight into the behavior and robustness of the NN-VFA under varying conditions. Third, we report the computational times corresponding to various VFA configurations. Fourth, we evaluate the performance (i.e., the direct rewards) of the tested VFAs.  We discuss offline performances over time -- fixing the policy after every 10,000 training iterations -- which is valuable when computational budgets are limited. We perform $N=100,000$ training iterations and 10,000 performance iterations per offline policy.

% Sutton:instability VFA
%https://papers.nips.cc/paper/3626-a-convergent-on-temporal-difference-algorithm-for-off-policy-learning-with-linear-function-approximation.pdf

All procedures are coded in C++ and CPLEX 12.8 is used to solve the linear decision problems. The experiments were run on a 64-bit Linux machine with a 4x1.60GHz CPU and 8GB RAM.

\subsection{Problem definition}\label{ssec:problemdefinitions}
This section outlines the transportation problem, which is based on the nomadic trucker problem \cite{powell2007}. It is characterized by a large discrete action space and a complex optimal policy. Let a strongly connected graph $\{\mathcal{V}, \mathcal{E}\}$ represent a transport network. Vertex set $\mathcal{V}$ represents the potential origins and destinations of transport jobs. Edge set $\mathcal{E}$ specifies the undirected connections between vertices. Each edge has travel time 1. Edge lengths are $L^2$ distances between vertex pairs, used to compute travel costs. A capacitated agent roams the graph, traveling between directly connected vertices. At each decision epoch $t \in \mathcal{T}$, the agent decides (i) which jobs to load, (ii) which jobs to unload, (iii) which vertex to visit next (or to stay at the current vertex). 

We sketch the corresponding MDP. The problem state $S$ contains the information necessary for decision-making, namely the relevant properties of all transport jobs in the graph and the current location of the agent $v^{loc}\in\mathcal{V}$. Each job is defined by four properties, namely (i) the vertex $v \in \mathcal{V}$ at which the job is currently located, (ii) the destination vertex $v^+ \in \mathcal{V}$, (iii) the time remaining until the due date $t^{+} \in \mathcal{T}^{+}$, and (iv) the assignment status $a\in\{0,1\}$ ($a=1$ means the job is currently carried by the agent). Each unique combination of properties constitutes a job type $[v,v^+,t^+,a]$; the number of jobs per type is denoted by $I_{v,v^+,t^{+},a}$. For the full system we define the vector
$I=[I_{v,v^+,t^{+},a}]_{\forall [v,v^+,t^{+},a]}$. The problem state is given by $S\triangleq[I,v^{loc}]$; the set containing all possible states is denoted by $\mathcal{S}$.

We proceed to describe the action $x$. Let $\mathcal{V}_{v^{loc}}^{adj} \subseteq \mathcal{V}$ be the set containing both $v^{loc}$ and the vertices adjacent to it. The variable $v^{nxt}\in \mathcal{V}_{v^{loc}}^{adj}$ describes the next destination of the agent. Furthermore, $\gamma=0$ indicates that a job is unloaded and $\gamma=1$ indicates that a job is loaded. The action is defined by
$x(S)=[x_{v,v^+,v^{nxt},t^{+},a,\gamma}]_{\forall [v,v^+,v^{nxt},t^{+},a,\gamma]}$.
%$x(S)=[x_{v,v^+,v^{nxt},t^{+},a,\gamma}]_{\forall [v,v^+,v^{nxt},t^{+},a,\gamma] \in \mathcal{V} \times \mathcal{V} \times \mathcal{V}_{v^{loc}}^{adj}   \times \mathcal{T}^{+}  \times  \{0,1\} \times  \{0,1\}}$. 
The action space $\mathcal{X}(S)$ is bound to various constraints; due to space limitations the full LP model is omitted. The key constraints are that (i) the agent may only (un)load at its current location, (ii) jobs are always unloaded when at their destination vertex, (iii) the agent's transport capacity may not be exceeded. 

%\textbf{Transition}\\
%The transition from state $S$ to $S^\prime$ depends on action $x$ and the stochastic realization of jobs generated between decision epochs. The state-action pair is computed deterministically. Subsequently, we take a time step to the next decision epoch, in which time indices $t^+$ are decremented and the agent location is updated from $v^{loc}$ to $v^{nxt}$. Jobs with $t^+=0$ are removed from the system. Finally, a sample of jobs is randomly generated in accordance with probability distribution $\mathbb{P}(S^\prime|S,x)$. When generated, it is ensured that the due date $t^+$ suffices to deliver the job via its shortest path. In addition, the job has a random amount of time slack. 

Next, we describe the reward function $R(S,x)$. The rewards consist of the following components: (i) a fixed reward for each successful delivery and (ii) a reward for bringing a job closer to its destination, proportional to the reduction in shortest path distance (an increase in distance yields a negative reward). We proceed to discuss the costs: (i) a fixed cost per distance unit covered (i.e., a cost associated with each edge), independent of the number of jobs carried, (ii) a fixed cost associated with each job that is (un)loaded, and (iii) a penalty for violating due dates. Jobs may be voluntarily unloaded by the agent or forced to be unloaded when $t^{+}=0$, i.e., when the due date has been reached. The reward components are linear with respect to jobs and distances, as to not give NN-VFAs an unfair advantage.
%Mathematically, the reward function is defined by

%\begin{align}
%R(S,x)=\sum_{\substack{[v,v^+,t^{+},a]\in \\ \mathcal{V} \times \mathcal{V}  \times \mathcal{T}^{+}\times \{0,1\}}} \bigg(r^{dst}(v,v^{nxt},v^+) \cdot (I_{v,v^+,t^{+},a}+ \notag\\
%x_{v,v^+,t^{+},a,1}-x_{v,v^+,v^{nxt},t^{+},a,0}) + \notag\\
%\bold{1}_{t^{+}>0,v^{loc}\neq v^{nxt}}(t^{+},v^{loc}) \cdot  r^{fst}(t^{+}) \cdot x_{v,v^+,v^{nxt},t^{+},a,1}\bigg) \notag\\
%-\sum_{\gamma \in \{0,1\}} x_{v,v^+,v^{nxt},t^{+},a,\gamma} \cdot c^l -  c(v^{loc},v^{nxt})\enspace. \notag
%\end{align}

%The features used in the experiments are inspired by \cite{vanheeswijk2018} and \cite{vanheeswijk2019}, reflecting the components of the reward function. 

Feature design reflects the components of the reward function. The features are low-level and expressed by linear equations based on $[S,x]$. We define the following features: a bias scalar, the number of jobs carried by the agent, the location of the agent, the number of jobs per vertex in $\mathcal{V}^{adj}$, the total time slack per neighboring vertex, and the most likely vertex to visit after visiting the neighboring vertex (given the shortest path of each job). The total number of features is $2+3\cdot|\mathcal{V}|$. %Table \ref{table:featurestransportproblem} summarizes the features.

For the experiments, we use an instance with $|\mathcal{V}|=5$, a maximum degree of 3, and up to 5 new jobs generated per vertex at each epoch, with accumulation possible up to 45 jobs. The agent may carry up to 20 jobs. The action space grows exponentially with the number of jobs, rendering enumeration infeasible even for this modest instance. An upper bound for the size of $|\mathcal{X}|$ is $\max(\mathcal{X}|)=2^{20}\cdot 2^{45} \cdot (3+1)$.

%\begin{table}[H]
%	\begin{tabular}{lll}
%		\textbf{Description}		& \textbf{\# features} & \textbf{\# nonzero features} \\
%		\hline
%		Constant & $1$	&   $1$ \\
%		\# jobs carried by agent	& $1$ & $1$  \\
%		Next agent location	& $|\mathcal{V}|$ & $1$ \\
%		\# jobs unassigned	& $|\mathcal{V}|$ & $|\mathcal{V}^{adj}|$ \\
%		Total slack per vertex	& $|\mathcal{V}|$ & $|\mathcal{V}_{v}^{adj}|$ \\
	%		Most likely vertex	& $|\mathcal{V}|$ & $|\mathcal{V}_{v}^{adj}|$ \\
%		\hline
%		Total	& $2+3\cdot|\mathcal{V}|$ & $3+2\cdot|\mathcal{V}^{adj}|$ \\
%		\hline
%	\end{tabular}
%	\label{table:featurestransportproblem}
%\end{table}

\section{Numerical results}\label{ssec:results}
\begin{figure}[h!]
\includegraphics{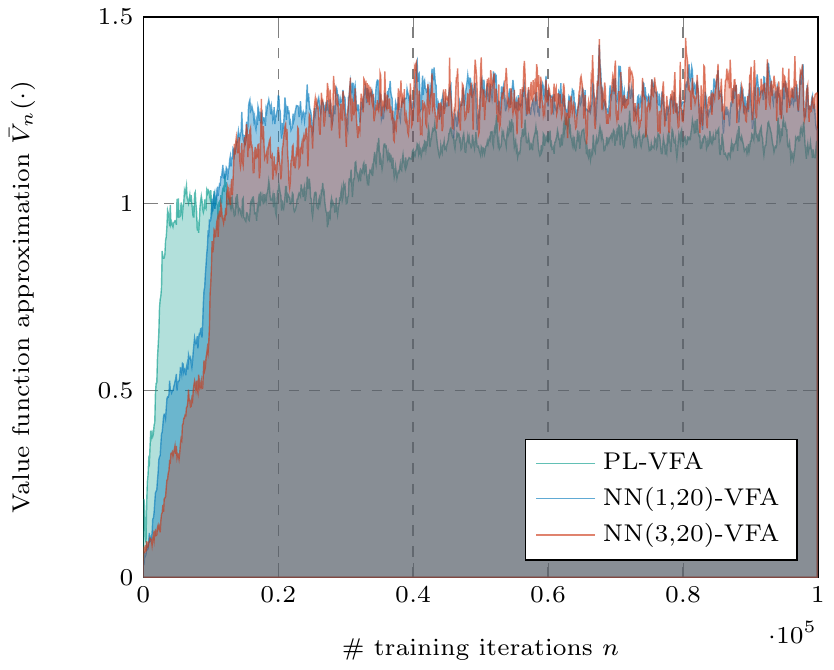}
\caption{Example of $\bar{V}_n(\cdot)$ convergence for various VFAs.}
\label{fig:convergencenetworks}
\end{figure}

This section discusses the results of the experiments. We start with the convergence results. Preliminary experiments on simplified problem settings with trivial policies indicate that all VFAs work correctly, converging to the true optimal value function, i.e., $\bar{V}_N(\cdot)\approx V(\cdot)$. Figure~\ref{fig:convergencenetworks} shows a convergence example for the real problem instance. The PL-VFA converges fastest, but to considerably lower values than the NN-VFAs. Similarly, the NN(1,20)-VFA converges faster than the NN(3,20)-VFA, but to somewhat lower values.

\begin{figure}[htb]
	\includegraphics{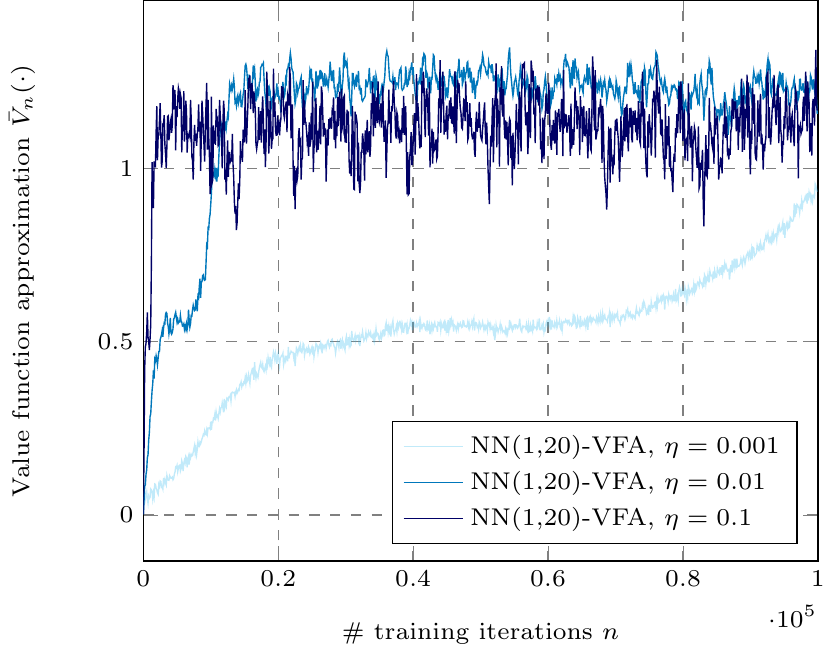}
	\caption{$\bar{V}_n(\cdot)$ for NN(1,20), using various learning rates $\eta$.}
	\label{fig:convergencelearningrates}
\end{figure}

The next experiment addresses learning rates, testing $\eta=\{0.001,0.01,0.1\}$. Figure~\ref{fig:convergencelearningrates} illustrates the convergence speeds per learning rate for the NN(1,20)-VFA; to aid the visual representation, we omit the other VFAs (which display comparable behavior). However, the NN(3,20)-VFA with $\eta=0.1$ did not converge to a stable policy. In general, we find that deeper neural networks are less robust with respect to larger learning rates. Errors may be magnified when passing through multiple layers, returning extreme values. Furthermore, NN-VFAs with $\eta=0.001$ do not converge within 100,000 iterations. We therefore use $\eta=0.01$ onwards.

Next, we look at the effects of altering network configurations, varying the number of neurons per layer. The results are shown in Table~\ref{table:networkconfigurations}. The results are fairly robust, with the exception of the NN(1,10)-VFA, which performs comparatively poorly. Balancing performance and speed, we use 20 neurons per layer for the remainder of the experiments. 

\begin{table}[htb]
\begin{tabular}{lrrrrr}
	\toprule
		&	\multicolumn{5}{c}{\textbf{\# neurons per layer}} \\
	\textbf{VFA} & \textbf{10}  &  \textbf{15} & \textbf{20}  & \textbf{25}  & \textbf{30} \\
	\midrule
	NN(1,$\cdot$)-VFA 	& 0.66 	& 0.91  & 1.00 & 0.92 & 0.98\\
	NN(3,$\cdot$)-VFA 	& 0.92  & 0.99  & 0.99 & 0.98 & 1.00  \\
	\bottomrule
\end{tabular}
\caption{Policy performance $R(\cdot)$ for the NN(1,$\cdot$)-VFA and NN(3,$\cdot$)-VFA with various \# neurons, normalized w.r.t. best VFAs.}
\label{table:networkconfigurations}
\end{table}
We assess the computational time per iteration; roughly 99\% of the computational budget is allocated to solving the LPs. On average, the polynomial VFA is solved in 0.02$s$ per iteration, the NN(1,20)-VFA takes 0.16$s$, and the NN(3,20)-VFA takes 0.39$s$. Due to the additional sets of variables and constraints, NN-VFAs are inherently slower to compute than the PL-VFA. Table~\ref{table:computationaltimes} shows the times for other network configurations also; both adding layers and neurons significantly increases the computational effort.
\begin{table}[htb]
\begin{tabular}{lrrrrrr}
	\toprule
&&	\multicolumn{5}{c}{\textbf{\# neurons per layer}} \\
\textbf{VFA}	&& \textbf{10}  &  \textbf{15} & \textbf{20}  & \textbf{25}  & \textbf{30} \\
	\midrule
	PL-VFA 	& 0.02	& -&   - & - &- & -\\
	NN(1,$\cdot$)-VFA	&-& 0.07	& 0.15 & 0.16 & 0.17 & 0.19\\
	NN(3,$\cdot$)-VFA 	&-& 0.15  & 0.33 & 0.39 & 0.66 & 0.71 \\
	\bottomrule
\end{tabular}
\caption{Computational time (in $s$) per iteration for various VFAs.}
\label{table:computationaltimes}
\end{table}

\begin{figure}[htb]
\includegraphics{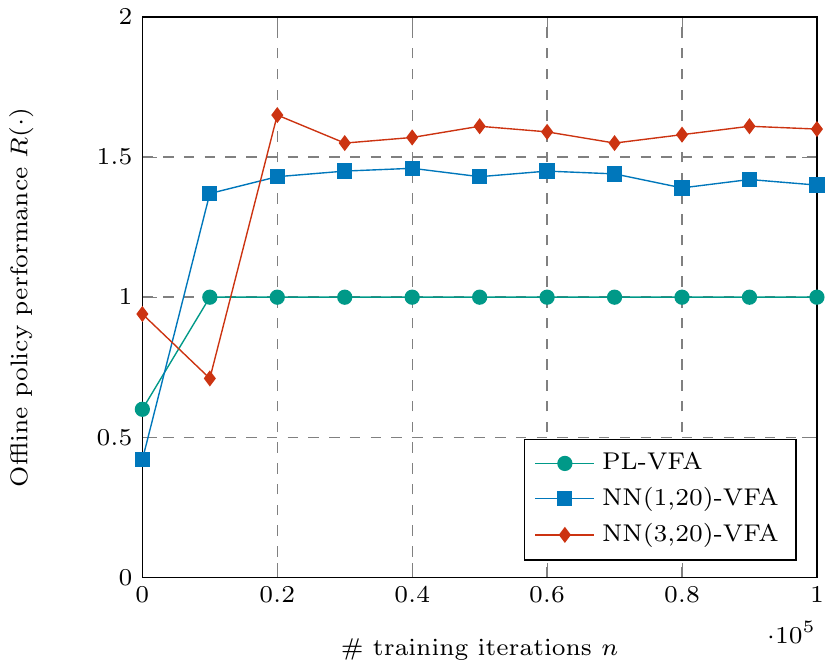}
\caption{Example of offline policy performance $R(\cdot)$ for various $\bar{V}_n, n \in \{0,N\}$}
\label{fig:offlineperformance}
\end{figure}

To conclude, we reflect on the qualities of the NN-VFA policies. An example of offline performances -- measured after each 10,000 training iterations -- is shown in Figure~\ref{fig:offlineperformance}. This example illustrates that from 20,000 iterations onwards, the PL-VFA is considerably outperformed. In general, we noted that the PL-VFA rather quickly results in a stable -- but often clearly suboptimal -- policy. Furthermore, the NN(3,20)-VFA performs better than the NN(1,20)-VFA and is more stable over time.

Table~\ref{table:averageperformance} shows the average policy performance of repeated replication, measured after completing $N=100,000$ training iterations. The NN(1,20)-VFA outperforms the PL-VFA by 10.1\% and the NN(3,20)-VFA does so by 19.6\%. Although both NN-VFAs achieve comparable policy performance at times, the NN(1,20)-VFA is more prone to fluctuating performances, which is also indicated by its higher standard deviation. Furthermore, the NN(3,20)-VFA simply has more expressive power. The results demonstrate that the NN-VFAs significantly outperform the PL-VFA for our transportation problem. 
\begin{table}[htb]
\begin{tabular}{lrr}
	\toprule
	\textbf{VFA}	& \textbf{Mean} &  \textbf{St. dev.}\\
	\midrule
	PL-VFA & $1.000$	&   $1.25\%$  \\
	NN(1,20)-VFA & $1.101$	&   $1.65\%$ \\
	NN(3,20)-VFA	& $1.196$ & $0.72\%$  \\
	\bottomrule
\end{tabular}
\caption{Average policy performance $R(\cdot)$ after $N$ iterations, normalized w.r.t. PL-VFA.}
\label{table:averageperformance}
\end{table}
\section{Conclusions}
This paper introduces the integration of linear programs and value function approximations in the form of neural networks, geared towards solving high-dimensional and combinatorial problems in operations research. Our proposed hybrid method is rooted in the framework of approximate dynamic programming. Traditionally, large action spaces in OR problems are handled by formulating the decision problem as a linear program, yet it is difficult to properly define polynomial VFAs in this context.   

The main contribution of the NN-VFA is the reduced effort of manual feature design, which is a crucial and precarious step in all solutions relying on VFAs. Unlike PL-VFAs, the NN-VFA is able to learn higher-order effects of simple input features without explicitly designing them, reducing the effort for manual feature design. This is particularly relevant when embedding VFAs in linear programs, in which the design of nonlinear features may be a cumbersome task.

We test our solution method on a representative transportation problem with a large discrete action space, a complex optimal policy, and a multi-component reward function. We compare NN-VFAs to the traditional PL-VFA, keeping all other factors equal. We observe significant improvements in performance. The findings are also robust with respect to neural network configurations; with various settings for training iterations, learning rates, neurons, and layers, the PL-VFA is consistently outperformed. NN-VFAs with multiple hidden layers yield the best and most stable policies, but also require more iterations to converge and more computational effort per iteration. We emphasize that this paper is an exploration of integrating LPs and NN-VFAs; additional research on different problems is needed to draw more general conclusions about the NN-VFA. In our opinion, the obtained results warrant such further studies.

\section*{Acknowledgments}
This work is part of the research program Scalable Interoperability in Information Systems for Agile Supply Chains (SIISASC) with project number 438-13-603, which is partially funded by the Netherlands Organization for Scientific Research (NWO).

%\newpage

%\newpage
\bibliographystyle{named}
\bibliography{references}

\begin{thebibliography}{}

\bibitem[\protect\citeauthoryear{Bertsekas and
  Tsitsiklis}{1995}]{bertsekas1995}
Dimitri Bertsekas and John Tsitsiklis.
\newblock Neuro-dynamic programming: an overview.
\newblock In {\em Proceedings of the 34th IEEE Conference on Decision and
  Control}, volume~1, pages 560--564. IEEE Publ. Piscataway, NJ, 1995.

\bibitem[\protect\citeauthoryear{Bertsekas}{2008}]{bertsekas2008}
Dimitri Bertsekas.
\newblock Neuro-dynamic programming.
\newblock In {\em Encyclopedia of optimization}, pages 2555--2560. Springer,
  2008.

\bibitem[\protect\citeauthoryear{Bunel \bgroup \em et al.\egroup
  }{2018}]{bunel2018}
Rudy~R Bunel, Ilker Turkaslan, Philip Torr, Pushmeet Kohli, and Pawan~K
  Mudigonda.
\newblock A unified view of piecewise linear neural network verification.
\newblock In {\em Advances in Neural Information Processing Systems}, pages
  4791--4800, 2018.

\bibitem[\protect\citeauthoryear{Cybenko}{1989}]{cybenko1989}
George Cybenko.
\newblock Approximation by superpositions of a sigmoidal function.
\newblock {\em Mathematics of Control, Signals and Systems}, 2(4):303--314,
  1989.

\bibitem[\protect\citeauthoryear{De~Farias and Van~Roy}{2003}]{defarias2003}
Daniela~Pucci De~Farias and Benjamin Van~Roy.
\newblock The linear programming approach to approximate dynamic programming.
\newblock {\em Operations Research}, 51(6):850--865, 2003.

\bibitem[\protect\citeauthoryear{Delalleau and Bengio}{2011}]{delalleau2011}
Olivier Delalleau and Yoshua Bengio.
\newblock Shallow vs. deep sum-product networks.
\newblock In {\em Advances in Neural Information Processing Systems}, pages
  666--674, 2011.

\bibitem[\protect\citeauthoryear{Dulac-Arnold \bgroup \em et al.\egroup
  }{2012}]{dulac2012}
Gabriel Dulac-Arnold, Ludovic Denoyer, Philippe Preux, and Patrick Gallinari.
\newblock Fast reinforcement learning with large action sets using
  error-correcting output codes for mdp factorization.
\newblock In {\em Joint European Conference on Machine Learning and Knowledge
  Discovery in Databases}, pages 180--194. Springer, 2012.

\bibitem[\protect\citeauthoryear{Gurney}{2014}]{gurney2014}
Kevin Gurney.
\newblock {\em An introduction to neural networks}.
\newblock CRC press, 2014.

\bibitem[\protect\citeauthoryear{Haykin}{2009}]{haykin2009}
Simon Haykin.
\newblock {\em Neural networks and learning machines}, volume~3.
\newblock Pearson Upper Saddle River, 2009.

\bibitem[\protect\citeauthoryear{He \bgroup \em et al.\egroup }{2015}]{he2015}
Kaiming He, Xiangyu Zhang, Shaoqing Ren, and Jian Sun.
\newblock Delving deep into rectifiers: Surpassing human-level performance on
  imagenet classification.
\newblock In {\em Proceedings of the IEEE international conference on computer
  vision}, pages 1026--1034, 2015.

\bibitem[\protect\citeauthoryear{Huang \bgroup \em et al.\egroup
  }{2016}]{huang2016}
Gao Huang, Yu~Sun, Zhuang Liu, Daniel Sedra, and Kilian~Q Weinberger.
\newblock Deep networks with stochastic depth.
\newblock In {\em European Conference on Computer Vision}, pages 646--661.
  Springer, 2016.

\bibitem[\protect\citeauthoryear{Lin \bgroup \em et al.\egroup
  }{2017}]{lin2017}
Henry Lin, Max Tegmark, and David Rolnick.
\newblock Why does deep and cheap learning work so well?
\newblock {\em Journal of Statistical Physics}, 168(6):1223--1247, 2017.

\bibitem[\protect\citeauthoryear{Pazis and Parr}{2011}]{pazis2011}
Jason Pazis and Ron Parr.
\newblock Generalized value functions for large action sets.
\newblock In {\em Proceedings of the 28th International Conference on Machine
  Learning (ICML-11)}, pages 1185--1192, 2011.

\bibitem[\protect\citeauthoryear{P{\'e}rez~Rivera and Mes}{2017}]{perez2017}
Arturo P{\'e}rez~Rivera and Martijn Mes.
\newblock Anticipatory freight selection in intermodal long-haul round-trips.
\newblock {\em Transportation Research Part E: Logistics and Transportation
  Review}, 105:176--194, 2017.

\bibitem[\protect\citeauthoryear{Powell \bgroup \em et al.\egroup
  }{2007}]{powell2007}
Warren~B Powell, Belgacem Bouzaiene-Ayari, and Hugo~P Simao.
\newblock Dynamic models for freight transportation.
\newblock {\em Handbooks in operations research and management science},
  14:285--365, 2007.

\bibitem[\protect\citeauthoryear{Powell}{2011}]{powell2011}
Warren Powell.
\newblock {\em Approximate Dynamic Programming: {S}olving the curses of
  dimensionality}, volume~2.
\newblock John Wiley \& Sons, 2011.

\bibitem[\protect\citeauthoryear{Powell}{2016}]{powell2016}
Warren Powell.
\newblock Perspectives of approximate dynamic programming.
\newblock {\em Annals of Operations Research}, 241(1-2):319--356, 2016.

\bibitem[\protect\citeauthoryear{Rolnick and Tegmark}{2018}]{rolnick2018}
David Rolnick and Max Tegmark.
\newblock The power of deeper networks for expressing natural functions.
\newblock In {\em International Conference on Learning Representations}, 2018.

\bibitem[\protect\citeauthoryear{Schmidhuber}{2015}]{schmidhuber2015}
J{\"u}rgen Schmidhuber.
\newblock Deep learning in neural networks: An overview.
\newblock {\em Neural networks}, 61:85--117, 2015.

\bibitem[\protect\citeauthoryear{Sutton and Barto}{2018}]{sutton2018}
Richard Sutton and Andrew Barto.
\newblock {\em Reinforcement learning: An introduction}.
\newblock MIT press, 2018.

\bibitem[\protect\citeauthoryear{Van~Heeswijk and
  La~Poutr{\'e}}{2018}]{vanheeswijk2018}
Wouter Van~Heeswijk and Han La~Poutr{\'e}.
\newblock Scalability and performance of decentralized planning in flexible
  transport networks.
\newblock In {\em 2018 IEEE International Conference on Systems, Man, and
  Cybernetics}, pages 292--297. IEEE, 2018.

\bibitem[\protect\citeauthoryear{Van~Heeswijk \bgroup \em et al.\egroup
  }{2019}]{vanheeswijk2019}
Wouter Van~Heeswijk, Martijn Mes, and Marco Schutten.
\newblock The delivery dispatching problem with time windows for urban
  consolidation centers.
\newblock {\em Transportation Science}, 53(1):203--221, 2019.

\bibitem[\protect\citeauthoryear{Wilmanski \bgroup \em et al.\egroup
  }{2016}]{wilmanski2016}
Michael Wilmanski, Chris Kreucher, and Jim Lauer.
\newblock Modern approaches in deep learning for {SAR} {ATR}.
\newblock In {\em Algorithms for synthetic aperture radar imagery XXIII},
  volume 9843. International Society for Optics and Photonics, 2016.

\end{thebibliography}

\end{document}